\documentclass[10pt,twocolumn,letterpaper]{article}

\usepackage{cvpr}              

\usepackage{url}
\usepackage{algorithm}
\usepackage{algpseudocode}
\usepackage{caption} 
\usepackage{graphicx}
\usepackage{booktabs} 

\definecolor{cvprblue}{rgb}{0.21,0.49,0.74}
\usepackage[pagebackref,breaklinks,colorlinks,allcolors=cvprblue]{hyperref}

\def\paperID{9208} 
\def\confName{CVPR}
\def\confYear{2025}

\title{Adaptive Consensus Gradients Aggregation for Scaled Distributed Training}
\author{Yoni Choukroun \quad\quad\quad Shlomi Azoulay \quad\quad\quad Pavel Kisilev\\
Huawei Tel Aviv Research Center \\
}

\begin{document}
\maketitle

\begin{abstract}
Distributed machine learning has recently become a critical paradigm for training large models on vast datasets. 
We examine the stochastic optimization problem for deep learning within synchronous parallel computing environments under communication constraints.
While averaging distributed gradients is the most widely used method for gradient estimation, whether this is the optimal strategy remains an open question. 
In this work, we analyze the distributed gradient aggregation process through the lens of subspace optimization. 
By formulating the aggregation problem as an objective-aware subspace optimization problem, we derive an efficient weighting scheme for gradients, guided by subspace coefficients. 
We further introduce subspace momentum to accelerate convergence while maintaining statistical unbiasedness in the aggregation. 
Our method demonstrates improved performance over the ubiquitous gradient averaging on multiple MLPerf tasks while remaining extremely efficient in both communicational and computational complexity.
A sample implementation of the method is available at \url{https://github.com/yoniLc/AdaCons}.
\end{abstract}
\section{Introduction}
Distributed optimization is essential for training modern deep neural networks on large-scale datasets.
Distributed environments can utilize different methods to parallelize computations, such as data parallelism \citep{ben2019demystifying} and model (e.g., tensor, pipeline) parallelism \citep{shoeybi2019megatron}, each offering distinct benefits and suited to various applications \citep{goyal2017accurate,brown2020gpt3}.  
This work focuses on the challenge of efficient gradient aggregation within synchronous data parallelism where each worker processes a different subset of data.

Synchronous data parallelism evenly distributes subsets of the dataset among several compute nodes/workers. 
Each node computes its local gradient before their \emph{aggregation} into a central (master) model.
While this aggregation is generally done at every iteration, model averaging \citep{zinkevich2010parallelized,yu2019parallel}, which averages individual models trained over parallel workers, was developed to reduce communication overhead.

The most ubiquitous aggregation methods remain linear combinations of the descent directions, such as averaging \citep{polyak1992acceleration} and its proximal variants \citep{zhang2015deep}.
These aggregation methods are generally efficiently implemented using modern all-reduce strategies \cite{chan2007collective}.
However, finding optimal aggregation remains an open problem since distributed systems are vulnerable to computing errors from the workers \citep{blanchard2017machine} or to out-of-distribution data samples inducing bad local gradients. 
Recently, researchers have begun wondering whether model averaging is the best strategy in a distributed setting \citep{xiao2020averaging}, and have started exploring more elaborate aggregation schemes \citep{sun2017ensemble,ji2019learning,maleki2021scaling}. 
In particular, given multiple workers' directions, the aggregation remains an ill-posed problem that must be solved according to a prior metric of interest.

In this work, we propose an efficient linear subspace optimization perspective to the problem of gradient aggregation.
Beyond the conceptual novelty, we make three technical contributions:
(i) We first formulate the aggregation problem as a subspace optimization problem created upon the original optimization objective.
(ii) We propose a first-order approximation of the solution allowing the efficient closed-form formulation of the model update.
(iii) We further extend the solution with a momentum-based unbiased estimator based on the statistics of the previous subspace coefficients.

The benefits of the method are as follows.
(i) The method outperforms the standard averaging technique by substantial margins on multiple training tasks while remaining scalable with respect to the number of workers.
(ii) The method requires only low communication overhead and negligible computational cost.
(iii) The method does not require hyper-parameter tuning (e.g., learning rate) or modification of the standard data-parallel setting.
\section{Related Works}
In the distributed data-parallel deep learning optimization setting \citep{li2020pytorch}, the communication between the workers occurs after all the nodes complete processing their data batches.
Model or gradient averaging \citep{yu2019parallel,zinkevich2010parallelized}, which (periodically) averages individual models or gradients trained over parallel workers, is generally applied as the aggregation method of choice.  
Elastic averaging was proposed by \citep{zhang2015deep} to allow local workers to fluctuate and explore near the center variable. 
This scheme allows communication and computation overlap, which can potentially reduce communication delays as in Overlap SGD \citep{wang2020overlap}.
Instead of pulling each worker model towards the average model,  Leader SGD \citep{teng2019leader} proposes to pull each worker model towards the best-performing worker during each iteration. 
The Ensemble-Compression scheme \citep{sun2017ensemble} suggests forming an ensemble of models from all workers to be further distilled to the central worker. 
Another approach proposes to learn to aggregate by gradient descent by replacing rule-based aggregation with a learned meta-aggregator \citep{ji2019learning}.
In the field of federated learning \citep{mcmahan2017communication}, model averaging is shown as suboptimal \citep{xiao2020averaging}.
{Adasum} \citep{maleki2021scaling} proposed an adaptive gradient aggregation simulating multiple sequential gradient descent steps.
Recently, \citep{dimlioglu2024grawa} proposed a new algorithm that periodically pulls workers towards the center variable computed as a weighted average of workers, where the weights are inversely proportional to the gradient norms of the workers in order to recover flat regions.

The idea of adaptive first-order optimization can be traced back at least to space dilation methods \citep{shor1983generalized}.
BackPROPagation \citep{braun1992rprop} originally proposed to scale the weights according to their sign. 
AdaGrad \citep{duchi2011adaptive} adjusted the learning rate based on the estimated geometry, assigning larger learning rates to less frequent features.
RMSProp \citep{hinton2012neural} extended AdaGrad's capabilities via momentum. 
Adam \citep{kingma2014adam} further improved RMSProp by incorporating a running average of gradients and became one of the most predominant optimizers for deep learning. 
Numerous subsequent works have proposed variants of Adam \citep{dozat2016incorporating,shazeer2018adafactor,reddi2019convergence,loshchilov2017decoupled, zhuang2020adabelief, you2019large} or topology aware optimization for generalization \citep{foret2020sharpness}.
Classical second-order optimization algorithms pre-condition the gradient by adding curvature information \citep{broyden1970convergence,nesterov2006cubic,conn2000trust}, while many Hessian approximation or efficient computation methods have emerged over the years \citep{becker1988improving,amari1998natural,pascanu2013revisiting,martens2010deep,schaul2013no,pascanu2013revisiting,ba2017distributed,zhang2022eva,yao2021adahessian, liu2023sophia}. 
However, variable metric methods generally lack success in dethroning adaptive first-order methods for large-scale stochastic optimization \citep{bottou2018optimization}.

The core idea of subspace optimization is to perform the optimization of the objective function in a small subspace spanned by a set of directions obtained from an available oracle.
A direct benefit of subspace optimization is that the low-dimensional optimization task at every iteration can be addressed efficiently using heavier optimization tools such as second-order methods.
The subspace structure may vary depending on the chosen optimization technique.
Early methods proposed to extend the minimization to a $d$-dimensional subspace spanned by $d$ various previous directions, such as gradients, conjugate directions, previous outer iterations or Newton directions \citep{cragg1969study,miele1969study,dennis1987generalized,conn1996iterated}.
The Krylov descent method defines the subspace as a series of preconditioned gradients \citep{vinyals2012krylov}.
Related to Krylov subspaces, the Conjugate Gradient (CG) method \citep{hestenes1952methods} reduces the search space to the current gradient and the previous step. 
\citep{nemirovski1982orth} provided optimal worst-case complexity in the convex setting with the ORTH-method by spanning the subspace with three directions based on aggregated iterates.
The Sequential Subspace Optimization algorithm \citep{Narkiss-2005,ZibEladSPM,choukroun2020primal,choukroun2021efficient} extends ORTH by adding the previous search directions to emulate CG's manifold expansion property.

\section{Method}
We consider the minimization of a function $F(w)$ in a distributed synchronous parallel computing environment with $N\in \mathbb{N}$ workers and a master. 
We focus on the stochastic optimization problem of the following form
\begin{equation}
\begin{aligned}
    \label{eq:dist_setting_1}
    \min_{w\in \mathbb{R}^{d}} F(w) := \mathbb{E}_{\zeta}f(w; \zeta),
\end{aligned}    
\end{equation}
where $w$ denotes the model parameter to be estimated and $\zeta$ is a random variable that follows the data probability distribution.
We consider the global variable consensus optimization \citep{Boyd_alternating_2011} which defines the distributed empirical risk minimization as
\begin{equation}
\begin{aligned}
    \label{eq:dist_setting_2}
    \min_{w} \sum_{i=1}^{N}\mathbb{E}f(w; \zeta^{i})\approx \sum_{i=1}^{N}f(w; \mathcal{D}_{i}),
\end{aligned}    
\end{equation}
where each $\zeta_{i}$ follows the data distribution and its empirical estimate $\mathcal{D}_{i}\subset \mathcal{D}$ denoting a sampled batch from the dataset $\mathcal{D}$.
As an unbiased estimator, the ubiquitous distributed Stochastic Gradient Descent algorithm's iteration is defined as
\begin{equation}
\begin{aligned}
    \label{eq:sgd_setting}
    w_{t+1} &= w_{t}-\frac{\eta_{t}}{N}\sum_{i=1}^{N}\nabla_{w}f(w;\mathcal{D}_{i})|_{w=w_{t}}\\
    &:=w_{t}-\frac{\eta_{t}}{N}\sum_{i=1}^{N}g_{i}(w_{t})
\end{aligned}    
\end{equation}
with $\eta_{t}$ the step size at time $t$.

Given $\{g_{i}(w_{t})\}_{i=1}^{N}$ the set of gradients to be aggregated, we aim to find an aggregation scheme $\psi_{t}:\{g_{i}(w_{t})\}_{i=1}^{N} \rightarrow \mathbb{R}^{d}$ such that the descent iterate is defined as
\begin{equation}
\begin{aligned}
    \label{eq:sgd_setting2}
    w_{t+1} = w_{t}-{\eta_{t}}\psi_{t}\big(\{g_{i}(w_{t})\}_{i=1}^{N}\big)
\end{aligned}    
\end{equation}

\subsection{Aggregation as Subspace Optimization}
\label{sec:subspace_subsec}
Given $N$ workers' descent directions $\{g_{i}(w_{t})\}_{i=1}^{N}$ (e.g., gradients or local steps) obtained from the workers and a given iterate $w_{t}$, we propose to solve the following optimal linear aggregation objective defined as
\begin{equation}
\begin{aligned}
    \label{eq:adacons_1}
    \min_{\alpha \in \mathbb{R}^{N}} f\bigg(w_{t}+\sum_{i=1}^{N}g_{i}(w_{t})\alpha_{i}; \mathcal{D}\bigg)
    :=  f\big(w_{t}+P_{t}\alpha; \mathcal{D}\big),
\end{aligned}    
\end{equation}
with $P_{t}\in \mathbb{R}^{d\times N}$ denotes the matrix obtained via the concatenation of the $N$ descent directions.
This setting is reminiscent of subspace optimization methods \citep{conn1996iterated,Narkiss-2005} where $P_{t}$ defines the subspace spanned by the gradients' directions where the low dimensional optimization is performed.
Eq \ref{eq:adacons_1} defines the best linear aggregation scheme (weighted average) with respect to the optimization objective of interest. 
We note that SGD corresponds to the specific case of the uniform distribution with $\alpha_{i}=1/N, \forall i$.
Also, we note the subspace optimization paradigm requires a rich subspace \citep{nemirovski1982orth,conn1996iterated,Narkiss-2005} such that each worker's direction should be representative enough of a descent direction for fast convergence. 

While \emph{full} (i.e., until local convergence) subspace optimization presents appealing computational complexity \citep{zibulevsky2013speeding} and convergence properties \citep{nemirovski1982orth,Narkiss-2005} it cannot be applied to modern stochastic distributed settings for two main reasons.
With the computational cost of the subspace gradient being similar to its full parameters counterpart \citep{conn1996iterated}, one would rather directly optimize over the original space in the stochastic setting \citep{seboost}.
Most importantly, the use of variable metric methods (e.g., quasi-/Newton) \citep{nocedal1999numerical, nesterov2018lectures} over the subspace, which lie the core of the subspace methods efficiency, greatly lacks generalization capabilities in large-scale stochastic optimization \citep{bottou2018optimization}.
Therefore, since higher-order optimization techniques are ill-suited, we propose to restrict the method to efficient first-order methods adapted to the modern all-reduce strategies.
\subsection{Subspace First-Order Expansion}
We propose to approximate the best subspace direction via first-order Taylor approximation near the current working point $w_{t}$.
Thus, given $\alpha_{0}=\mathbf{0}$ and $\lambda \in \mathbb{R}_{+}$ the first-order approximation step size of the subspace objective, we have 
\begin{equation}
\begin{aligned}
    \label{eq:adacons_2}
    \alpha^{*}_{t} &= \alpha_{0} - \lambda \nabla_{\alpha} f(w_{t}+P_{t}\alpha;\mathcal{D})|_{\alpha=\alpha_{0}}
    \\
    &=-\lambda P^{T}\nabla_{w}f(w_{t};\mathcal{D})
\end{aligned}    
\end{equation}
Using Monte-Carlo approximation of the gradients \citep{mohamed2020monte} according to the workers' directions, we assume $\nabla_{w}f(w_{t};\mathcal{D})=\frac{1}{N}\sum_{i=1}^{N}g_{i}(w_{t}):=\bar{g}(w_{t})$.
Since the subspace optimization setting is invariant to the scaling of the spanning directions, we assume column normalized $P_{t}$ \citep{ZibEladSPM} such that we have $\forall i \in \{1,\dots, N\}$
\begin{equation}
\begin{aligned}
    \label{eq:adacons_3}
    (\alpha^{*}_{t})_{i} &= -\lambda\langle (P_{t})_{i}, \bar{g}(w_{t})\rangle 
    \\
    &= -{\lambda} \frac{\langle g_{i}(w_{t}), \bar{g}(w_{t})\rangle}{\|g_{i}(w_{t})\|} 
    \\
    &=-\frac{\lambda}{N}\sum_{j=1}^{N}   \frac{\langle g_{i}(w_{t}), g_{j}(w_{t})\rangle}{\|g_{i}(w_{t})\|}
\end{aligned}    
\end{equation}
Thus, the parameters' update is defined by the reprojection over the original space such that
\begin{equation}
\begin{aligned}
    \label{eq:adacons_4}
    w_{t+1} &=w_{t}-\lambda\frac{\eta_{t}}{N} P_{t}\alpha^{*}_{t} 
    \\
    &= w_{t}-\lambda\frac{\eta_{t}}{N} \sum_{i,j=1}^{N}   \frac{\langle g_{i}(w_{t}), g_{j}(w_{t})\rangle}{\|g_{i}(w_{t})\|^{2}}g_{i}(w_{t})
\end{aligned}    
\end{equation}

We can observe the subspace optimization formalism yields an \emph{adaptive} aggregation approach that enhances directions with the largest \emph{consensus}. The method is therefore dubbed as \emph{AdaCons}. 
Interestingly, this update is diametrically opposed in its effect to the existing sequential SGD approximation which enhances orthogonal directions \citep{maleki2021scaling} while our approach suggests the exploitation of similar directions.
Also, we note that assuming all the $g_{i}$s equal would reduce the proposed aggregation scheme to the standard gradient averaging formulation.
Finally, other optimizers (e.g., Adam \citep{kingma2014adam}) can be applied to the obtained aggregated directions.

\subsection{A Preconditioned Gradient Perspective}
The subspace optimization paradigm infers a preconditioned gradient method via the subspace matrix.
Given the preconditioning matrix $G$, the aggregated update can be written as
\begin{equation}
\begin{aligned}
    \label{eq:precond1}
    \psi_{t}\big(\{g_{i}(w_{t})\}_{i=1}^{N}\big) &= G \nabla_{w}f(w_{t};\mathcal{D}) 
    \\
    &:= \lambda P_{t}P^{T}_{t}\nabla_{w}f(w_{t};\mathcal{D}) 
    \\
    &= \lambda  \bigg(\sum_{i=1}^{N} \frac{g_{i}(w_{t}) g_{i}(w_{t})^{T}}{\| g_{i}(w_{t}) \|^{2}}\bigg) \nabla_{w}f(w_{t};\mathcal{D}) 
\end{aligned}    
\end{equation}
This emerging formulation is certainly evocative of the Natural gradient \citep{amari1998natural,pascanu2013revisiting} or Gauss-Newton \citep{nocedal1999numerical} methods where curvature information is integrated into the optimization. 
However, the Hessian approximation emerges in its non-inverted form through the subspace optimization framework.
As a sum of rank-one matrices, $G$ is positive-definite and ensures a descent direction.
\subsection{Adaptive Subspace Coefficients}
Momentum gradient descent, or the heavy ball method \citep{polyak1992acceleration} combines the current gradient with the history of the previous steps to accelerate the convergence of the algorithm is ubiquitous in modern optimization methods.
To avoid potential instabilities during training and enforce smoothness between consecutive subspace aggregations, we propose to apply the momentum method over the subspace coefficients, such that, with $\beta\in (0,1)$ we have
\begin{equation}
\begin{aligned}
    \label{eq:momentum_1}
     \alpha^{*}_{t} \leftarrow \alpha^{m}_{t} = \beta \alpha^{m}_{t-1} + (1-\beta)\alpha^{*}_{t}
\end{aligned}    
\end{equation}
with $\alpha^{m}_{t}$ the exponential moving average (EMA) of the subspace coefficients at iteration $t$.
However, each worker's batch is arbitrarily distributed, and each gradient's coefficient should be decoupled from its index.
Thus, to maintain the smooth subspace coefficients' \emph{distribution}  between consecutive iterations, we enforce invariance to the ordering by sorting the coefficients. Then the momentum moving average is performed to redistribute each smoothed coefficient to its corresponding worker.
Formally, given the sorting operator $\mathcal{S}$ we have
\begin{equation}
\begin{aligned}
    \label{eq:momentum_2}
    \alpha^{m}_{t} &= \beta \alpha^{m}_{t-1} + (1-\beta)\mathcal{S}(\alpha^{*}_{t})\\  
    \alpha^{*}_{t} &\leftarrow \mathcal{S}^{-1}(\alpha^{m}_{t})
\end{aligned}    
\end{equation}
This approach allows a smoother aggregation between iterations avoiding potentially unstable local gradients.

Finally, to enforce unbiased estimation, we further constrain the weighting coefficients to sum to one.
By rewriting the weighted aggregated direction more compactly, we have  
\begin{equation}
\begin{aligned}
    \label{eq:adacons_5}
    \psi_{t}\big(\{g_{i}(w_{t})\}_{i=1}^{N}\big) &:= {AdaCons}\big(\{g_{i}(w_{t})\}_{i=1}^{N}, t\big) \\
    &= \sum_{i}^{N} \gamma_{i}g_{i}(w_{t})
\end{aligned}    
\end{equation}
with $\gamma_{i}= (\alpha^{*}_{t})_{i}=\lambda g_{i}(w_{t})^{T}\bar{g}(w_{t})/\|g_{i}\|^{2}$ and where the unbiased estimator with respect to $P_{t}$ is given by enforcing sum one normalization such that 
\begin{equation}
\begin{aligned}
    \label{eq:adacons_6}
\lambda = 1/\sum_{i=1}^{N}\big( g_{i}(w_{t})^{T}\bar{g}(w_{t})/\|g_{i}\|\big )\end{aligned}    
\end{equation}
removing dependency on the $\lambda$ hyperparameter as a byproduct.  
Finally, as an unbiased estimator, the method \emph{inherits the convergence properties} of first-order optimizers \citep{bottou2018optimization}.
\subsection{Distributed System Implementation}
\label{sec:dist_sys}

\begin{figure}[t]
\centering
\includegraphics[trim={0 60 0 0},clip,width=1\columnwidth]{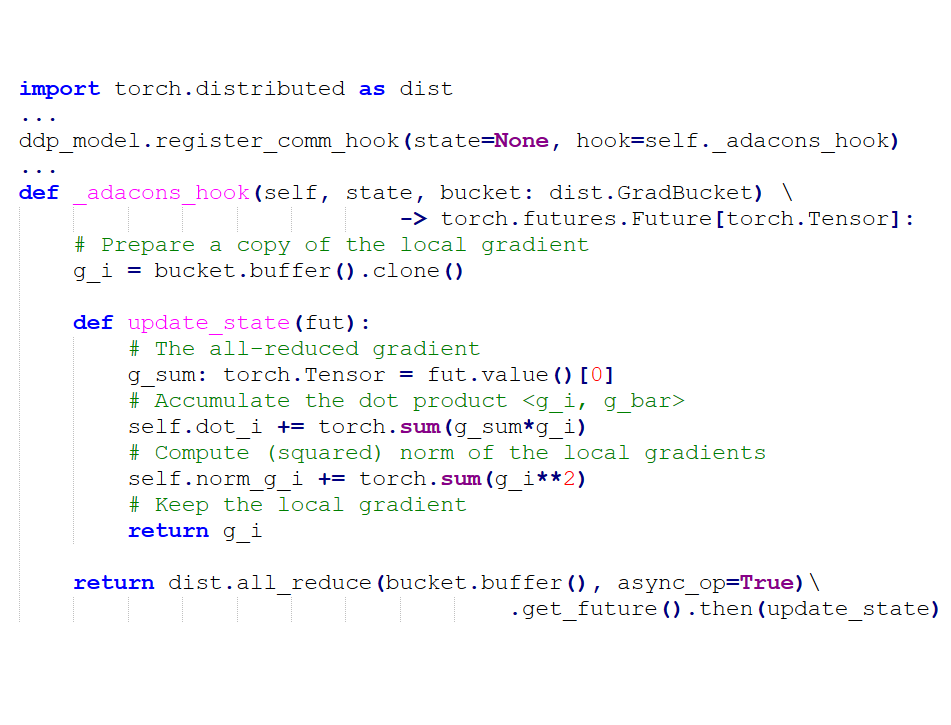}
\caption{Pytorch \citep{paszke2019pytorch} implementation of the AdaCons DDP communication hook.}
\label{fig:ddp_hook}
\end{figure}

We present in Algorithm \ref{algo:complexity} the communication and computational complexity of the method. 
Since the number of workers $N$ is negligible compared to the dimension $d$ of the model (i.e., $N\ll d$), the method only requires an additional asynchronous (ring) all-reduce of the weighted gradients compared to the traditional averaged aggregation. 

The method begins by aggregating the gradient and computing the subspace coefficients (Eq. \ref{eq:adacons_3}) using a single asynchronous all-reduce call.
After the coefficients are shared among the workers via all-gather, the subspace momentum (Eq. \ref{eq:momentum_2}) is applied followed by the unbiased estimation (Eq. \ref{eq:adacons_6}).
Finally, each worker locally adjusts its gradient using the corresponding weight factor, before performing a final all-reduce summation of the directions.

\begin{algorithm}
\caption{AdaCons Algorithm}\label{alg:cap}
\label{algo:complexity}
\begin{algorithmic}[1]
\State Compute $\alpha_{i}={g_{i}(w_{t})^{T}\bar{g}(w_{t})}$ via asynchronous all-reduce of $g_{i}$
\newline \Comment{$\mathcal{O}(d)$ Communication}
\State All-gather (broadcast) $\alpha_{i}/{\|g_{i}(w_{t})\|}$ \newline \Comment{$\mathcal{O}(N)$ Communication}
\State Compute normalized momentum $\gamma_{i}$  
\newline \Comment{$\mathcal{O}(N \log (N))$ Computation}
\State Compute final aggregation via the asynchronous all-reduce of $\gamma_{i}g_{i}(w_{t})$ 
\newline \Comment{$\mathcal{O}(d)$ Communication}
\end{algorithmic}
\end{algorithm}

AdaCons can be efficiently implemented using existing Distributed Data-Parallel (DDP) frameworks, with the most significant operation being the dot product computation (Eq. \ref{eq:adacons_3}). This computation can be seamlessly handled by leveraging existing DDP all-reduce hooks, as illustrated in Figure \ref{fig:ddp_hook} using PyTorch.

\begin{figure*}[t]
\centering
\noindent  \begin{tabular}{@{}ccc@{}}
        \includegraphics[trim={0 0 0 0},clip, width=0.31\linewidth]{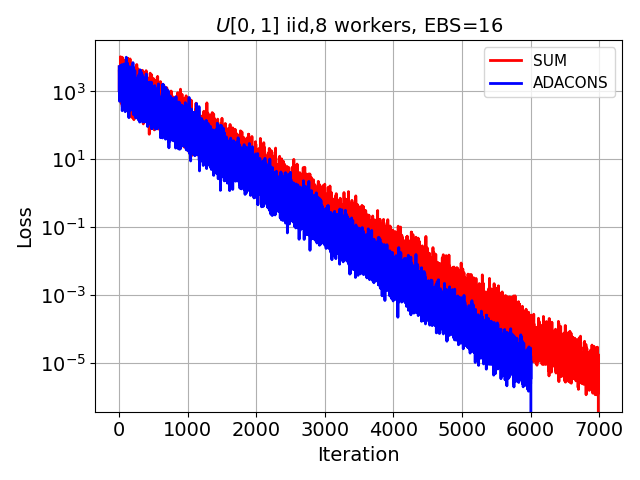}&
        \includegraphics[trim={0 0 0 0},clip, width=0.31\linewidth]{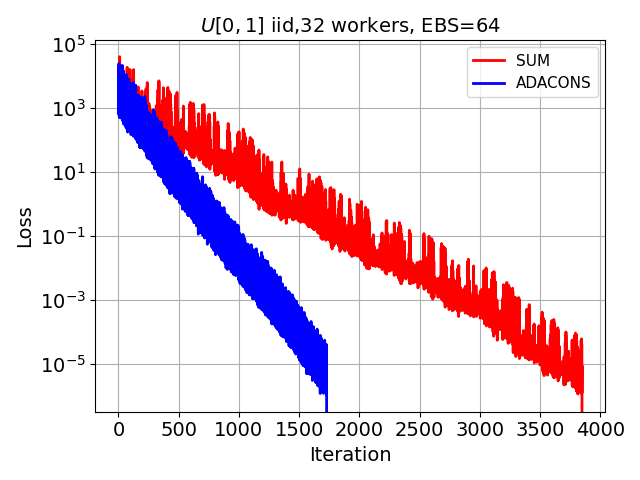}&
        \includegraphics[trim={0 0 0 0},clip, width=0.31\linewidth]{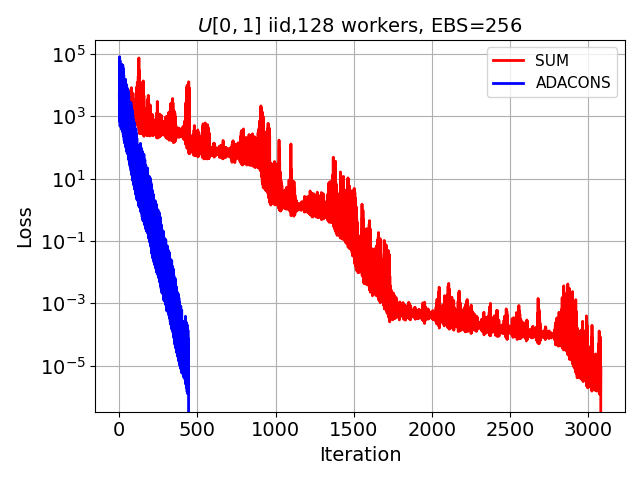}
        \\        
        \includegraphics[trim={0 0 0 0},clip, width=0.31\linewidth]{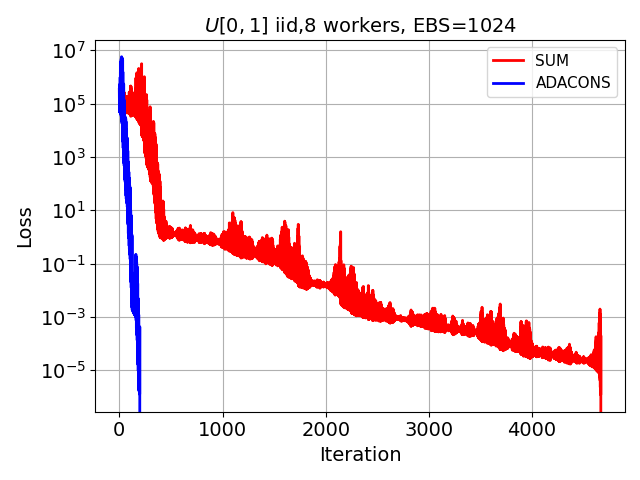}&
        \includegraphics[trim={0 0 0 0},clip, width=0.31\linewidth]{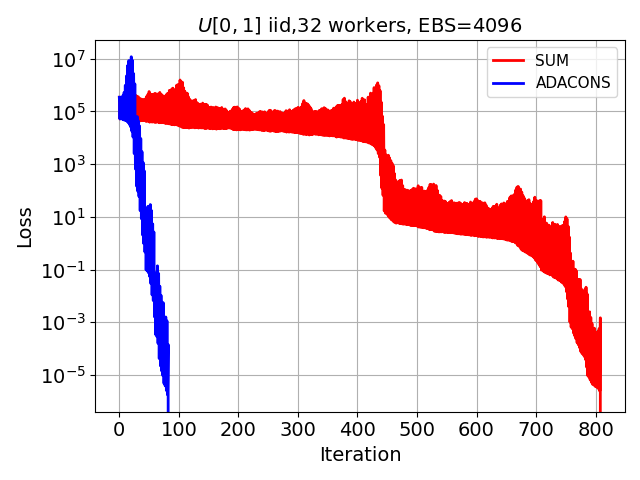}&
        \includegraphics[trim={0 0 0 0},clip, width=0.31\linewidth]{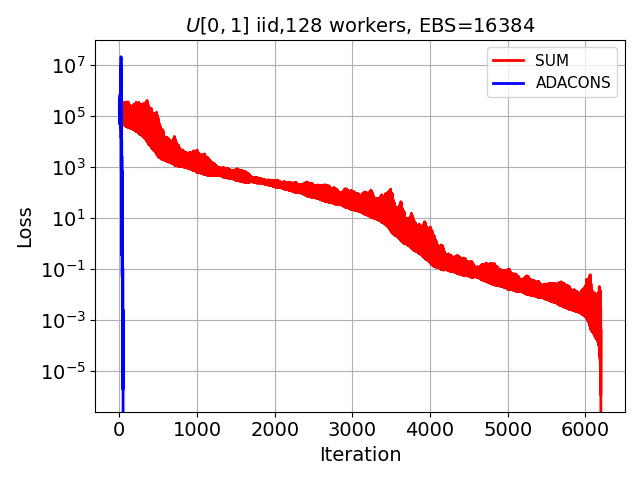}
        \\
      \end{tabular}
  \caption{Performance of the aggregation schemes on the stochastic linear regression tasks for various numbers of workers and effective batch sizes. 
  Additional experiments are given in Appendix \ref{appendix:linear}.
  }
\label{fig:linear}
\end{figure*}

\section{Experiments}
Due to the unbiased formulation of the algorithm, the proposed aggregation method offers a parameter-free framework that eliminates the need for hyper-parameter tuning (e.g., learning rate) or modifications to the aggregation setting (e.g., double precision accumulation or aggregation of the local optimizer updates \citep{maleki2021scaling}).
As a result, any performance improvements stem solely from the quality of the aggregated gradients, while the training setup remains identical to the baseline.
Moreover, as described in Section \ref{sec:dist_sys}, the method requires minimal code integration thanks to communication hooks that fully manage the aggregation process.

The distributed training infrastructure is constrained to a maximum of 8 nodes, each equipped with 4 RTX A6000 GPUs (48 GB per GPU), totaling 32 workers ($N=32$), connected via a 100 Gb/s Infiniband network. These resource limitations prevent us from scaling our experiments to very large tasks, as discussed in Section \ref{sec:limitations}.

Our analysis begins with simulated linear regression and then moves to four MLPerf V4.0 tasks \citep{mattson2020mlperf}, where we demonstrate the performance gains of our method, AdaCons, compared to standard baseline gradient summing/averaging (referred to as Sum). 
All the official baseline settings and implementations can be found in the official MLPerf repository \citep{mlperfrepo}. We remind the reader that our method is tested on the exact same training and evaluation settings and integrated into the same implementation.
The aggregation is computed model-wise, while layer-wise aggregation presents similar performance on the tested benchmark.
All batch sizes refer to the effective batch size, calculated as the local batch size multiplied by the number of workers.
Finally, we do not present results for the method of \cite{maleki2021scaling}, as we observed no improvement over the baseline (Sum), despite its increased computational complexity and distinct aggregation setting.

\begin{figure*}[t]
\centering
\noindent  \begin{tabular}{@{}ccc@{}}
        \includegraphics[trim={0 0 0 0},clip, width=0.32\linewidth]{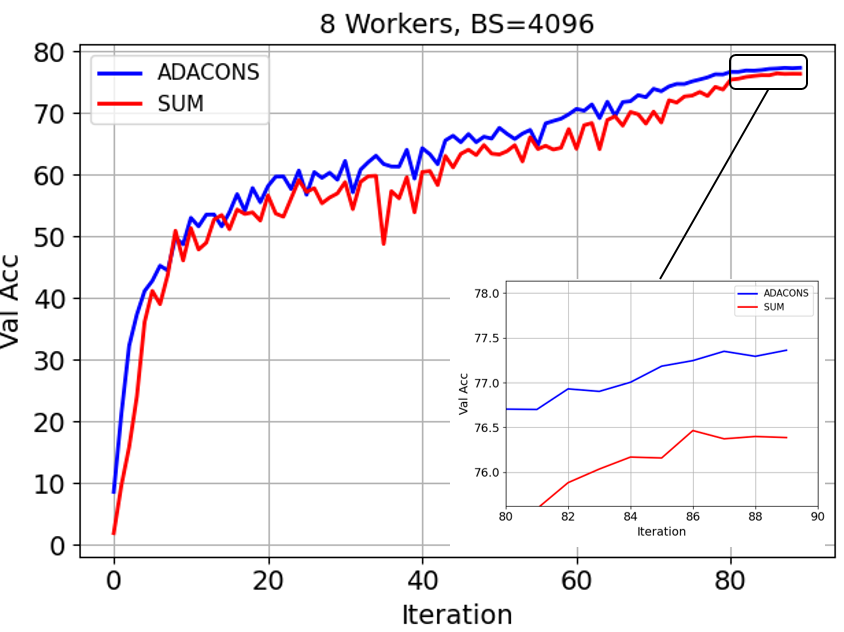} &
        \includegraphics[trim={0 0 0 0},clip, width=0.32\linewidth]{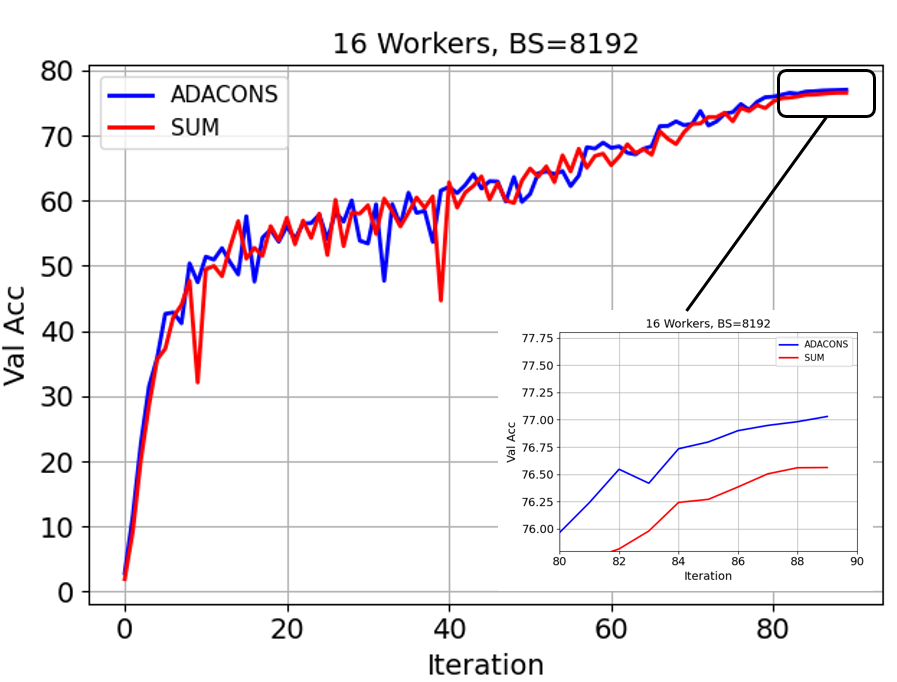} &
                \includegraphics[trim={0 0 0 0},clip, width=0.3\linewidth]{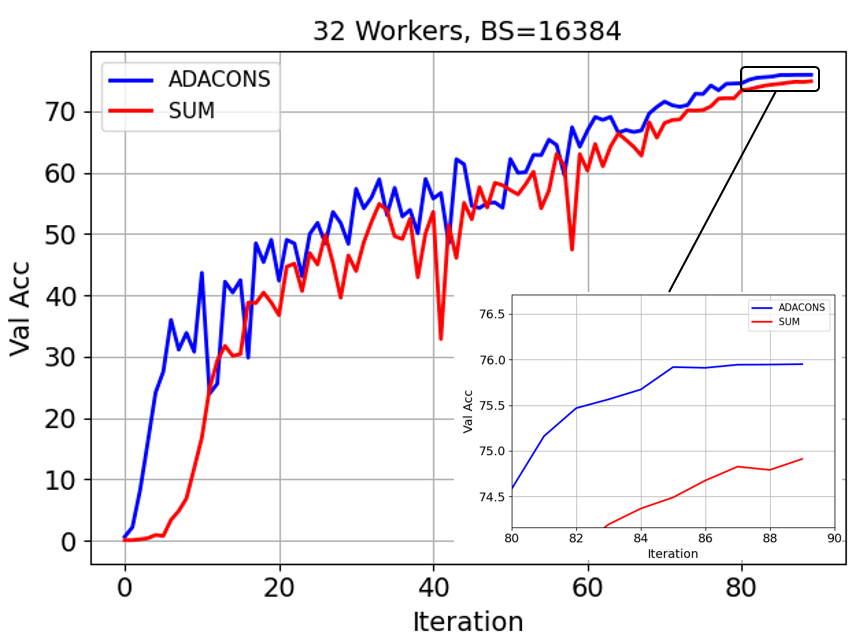}\\
      \end{tabular}
   \caption{Performance of the aggregation schemes on the MLPerf Imagenet classification task for various numbers of workers.  }
  \label{fig:imagenet}
\end{figure*}

\subsection{Stochastic Linear Regression}
We first analyze our method with the stochastic linear regression task where the optimization objective is given by 
\begin{equation}
\begin{aligned}
    \label{eq:linear}
    \min_{w\in \mathbb{R}^{1000}} \mathbb{E}_{\zeta\sim U[0,1]}\bigg[\frac{1}{2}(w^{T}\zeta)^{2}\bigg]
\end{aligned}    
\end{equation}
For a fair, hyperparameter-free comparison, we provide each method with the optimal (analytical) step size using SGD.
The results, presented in Figure \ref{fig:linear} analyze the impact of varying batch sizes and the number of workers. The performance improvement of our method is evident, particularly when using a large number of directions combined with sufficiently large batch sizes enabling richer subspace, which significantly enhances overall performance.

\subsection{Imagenet}
We present the performance of AdaCons on the MLPerf classification task using the ResNet-50v1 \citep{he2016deep} architecture on the ImageNet \citep{russakovsky2015imagenet} dataset, where the baseline performance is achieved with 8 workers. The results are shown in Figure \ref{fig:imagenet}.
Our method demonstrates improved convergence properties while maintaining scalability, resulting in a consistent improvement of $1\%$ in final accuracy.

\subsection{RetinaNet}
\begin{figure}[h]
\centering
\noindent  
\begin{tabular}{@{}ccc@{}}
        \includegraphics[trim={0 0 0 0},clip, width=0.9\linewidth]{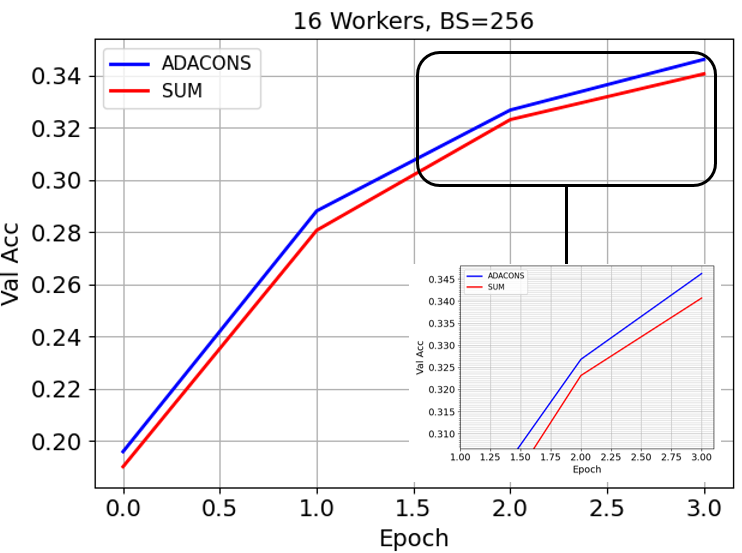}\\
        \includegraphics[trim={0 0 0 0},clip, width=0.9\linewidth]{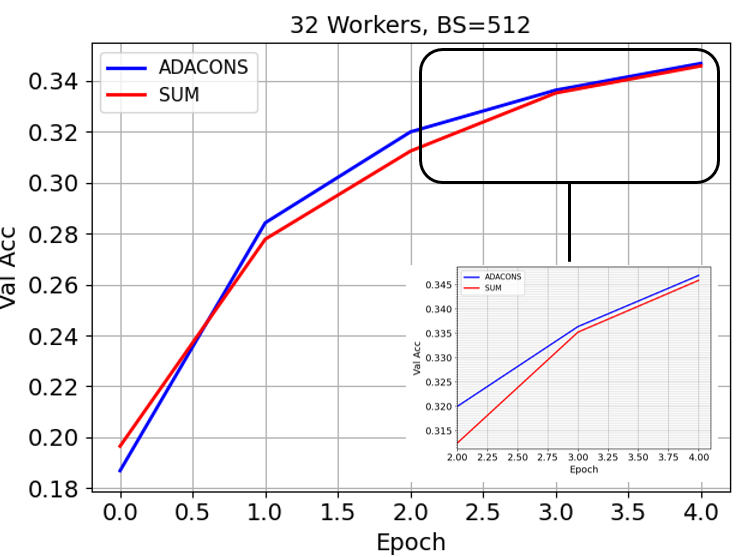}
      \end{tabular}
  \caption{Performance of the aggregation schemes on the MLPerf RetinaNet object detection task for various numbers of workers.
  }
  \label{fig:ssd}
\end{figure}

We evaluate the performance of AdaCons for the MLPerf object detection training task using RetinaNet \citep{lin2017focal} built upon the SSD architecture \citep{liu2016ssd}. The baseline performance is obtained using 16 workers (the target mAP is 0.34).
Figure  \ref{fig:ssd} illustrates the performance comparison.
We can observe that our method presents better convergence property while maintaining a positive $0.7\%$ and $0.2\%$ final accuracy (mAP) gap with 16 and 32 workers, respectively.

\subsection{Deep Learning Recommendation System}
We further show the performance of the proposed aggregator on the MLPerf Deep Learning Recommendation System (DLRM) \citep{naumov2019deep} training task defined the DCN V2 architecture \citep{wang2017deep}.
The baseline performance is given with a batch size of 64K workers with a target Area Under the Curve (AUC) of $0.8025$.
Due to the high memory requirement, the experiments of this task only have been performed with $8\times 80$GB A100 NVIDIA workers (i.e., $N=8$). 
We can observe that our method presents remarkable scaling properties and is able to surpass the target accuracy on up to $8\times$ scaling. 

\begin{figure*}[h]
\centering
\noindent  \begin{tabular}{@{}cccc@{}}
        \includegraphics[trim={0 0 0 0},clip, width=0.3\linewidth]{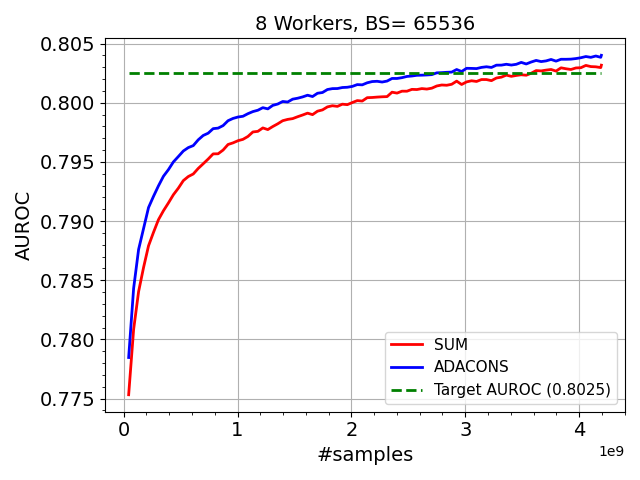} &
        \includegraphics[trim={0 0 0 0},clip, width=0.3\linewidth]{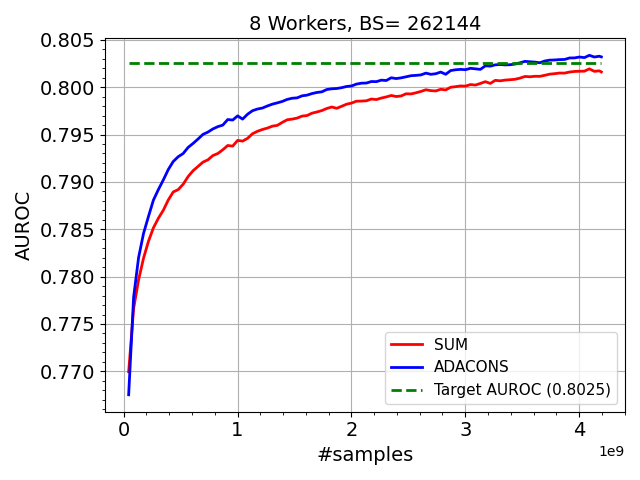} &
        \includegraphics[trim={0 0 0 0},clip, width=0.3\linewidth]{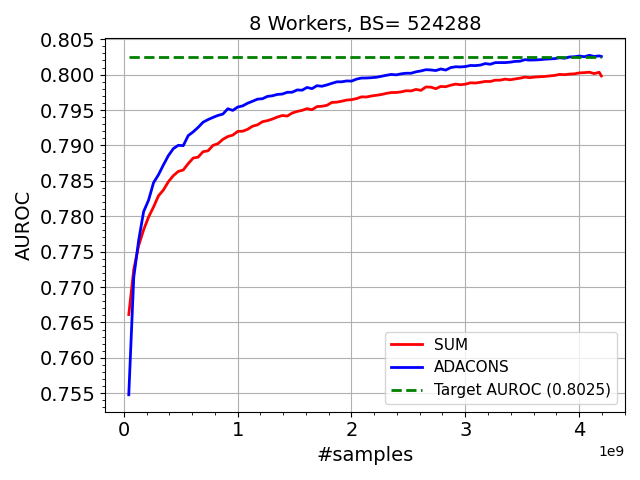} \\
      \end{tabular}
  \caption{Performance of the aggregation schemes on the MLPerf DLRM task for various batch sizes.
  Additional results and visualizations are given in Appendix  \ref{appendix:dlrm}.
  }
\label{fig:dlrm}
\end{figure*}

\subsection{BERT}
\label{sec:bert}

\begin{figure}[h]
\centering
\noindent  \begin{tabular}{@{}ccc@{}}
        \includegraphics[trim={0 0 0 0},clip, width=0.9\linewidth]{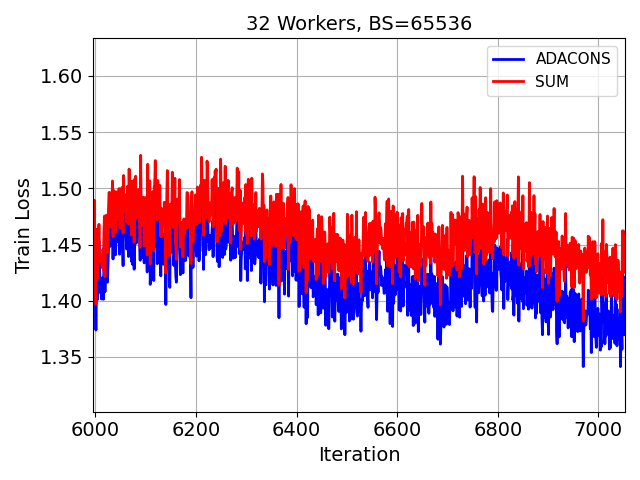}\\
        (a) \\
        \includegraphics[trim={0 0 0 0},clip, width=0.9\linewidth]{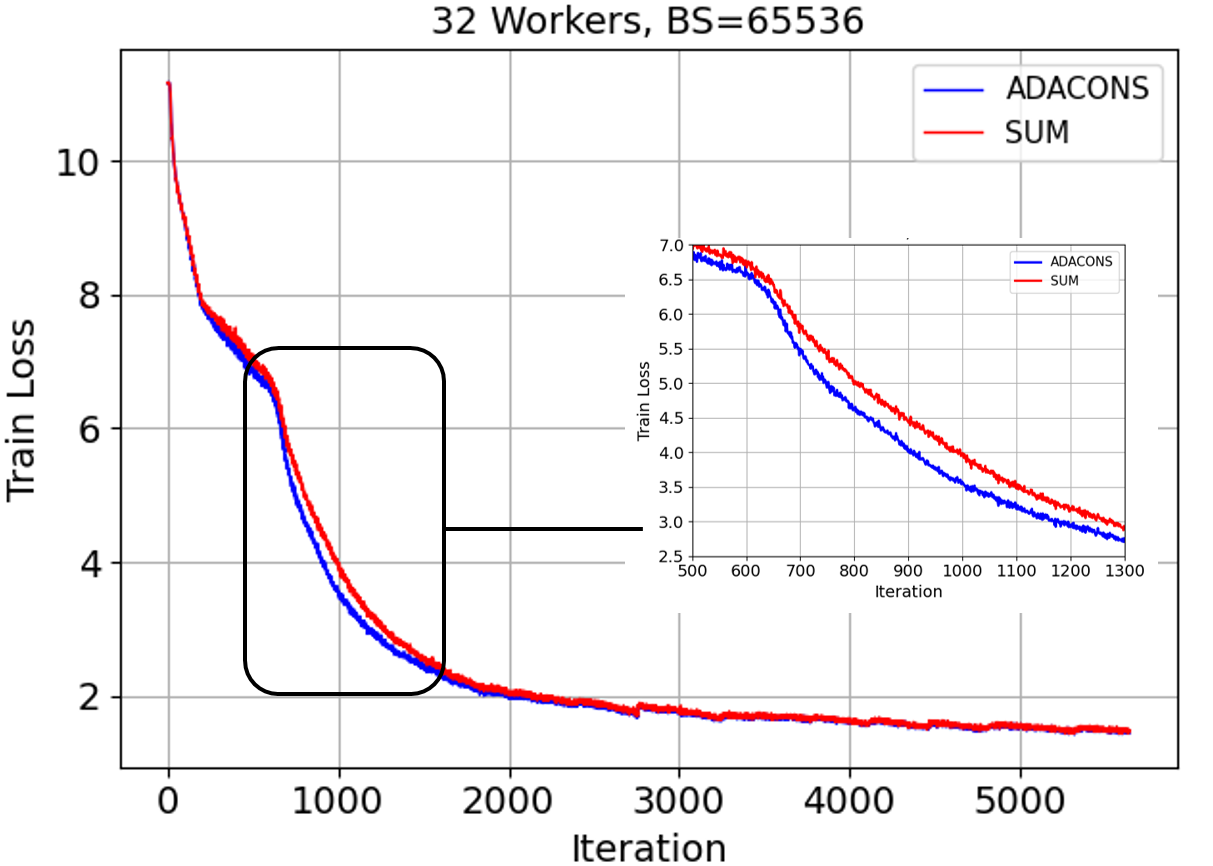}\\
        (b)
      \end{tabular}
  \caption{Performance of the aggregation schemes on the MLPerf BERT task on the baseline setting of interest. (a) standard setting (last 1K iterations), (b) 20\% less training iterations. Adacons minimal loss is (a) 1.34 (vs 1.38) and (b) 1.41 (vs 1.43).
  Additional visualizations are presented in Appendix \ref{appendix:bert}.
  }
\label{fig:bert}
\end{figure}

We present the performance of our framework during the pretraining (phase one) of BERT-Large \citep{devlin2018bert}. We evaluate the baseline setting (batch size = $64K$ with $7.037K$ iterations) as well as a 20\% reduction in iterations ($5K$) \citep{maleki2021scaling}. 
For the original setting, we observed a $3\%$ minimal accuracy gap (1.381 vs 1.341) with a $14\%$ speedup to reach the baseline minimum loss value.
For the second setting, a $1\%$ final accuracy gap only is observed with a $6\%$ speedup.
In both settings, a gap emerges during the initial stages of training, as shown in Figure \ref{fig:bert}. 
Section \ref{sec:limitations} discusses the possible reasons for these performance.

\section{Analysis}
In this section we present the analysis of the proposed method, including timing, ablation of the algorithmic components, and limitations study.
\subsection{Timing}
Table \ref{tab:timing} presents the per-iteration timing (in seconds) of the proposed method compared to the native PyTorch all-reduce baseline. We observe that the proposed method induces only a slight slowdown, ranging from $1.04\times$ to $1.05\times$. 
It is important to note that our implementation operates at a higher level of abstraction, leaving room for further optimizations. 
Furthermore, the communication setup used in our experiments is relatively slow compared to modern network speeds (800 Gb/s), where both the low-dimensional all-gather and the model all-reduce communications would become negligible.
\begin{table}[h]
    \centering
    \caption{
    Per iteration timing in seconds of the proposed method compared to the native all-reduce implementation on the different MLPerf tasks.
    }
    \label{tab:timing}
    \resizebox{0.99\columnwidth}{!}
    {%
    \begin{tabular}{l|cccc}
    \toprule
        Task & Imagenet & RetinaNet & DLRM & BERT  \\
        \midrule                                                              
		Sum           & $1.08\pm 0.272$ & $2.41\pm 0.51$ & $1.01\pm 0.13$ & $7.97\pm 0.02$ \\
		\midrule                                                                
    	AdaCons       & $1.11\pm 0.20$ & $2.51\pm 0.272$ & $1.07\pm 0.14$ & $8.28\pm 0.65$ \\
        \midrule                                                                
        \midrule                                                                
        Slowdown       & $1.04\times$ & $1.04\times$ & $1.05\times$ & $1.04\times$ \\
		\bottomrule
	\end{tabular}
	}
\end{table} 

\subsection{Ablation}
Table \ref{tab:ablation} presents the impact of different variants of the method on performance. 
We demonstrate the improvements brought by applying exponential moving average to the subspace coefficients and by normalizing the method (i.e., ensuring unbiasedness).
We observe that normalization leads to improved performance, likely due to better automatic scaling, while the addition of momentum further enhances the convergence.
\begin{table}[h]
    \centering
    \caption{
    Ablation study of the different variants of the method on the Imagenet (final validation accuracy) and BERT (final training loss) tasks.
    \emph{AdaCons} represent the basic subspace aggregation presented in Eq. \ref{eq:adacons_4} ($\lambda=1$), \emph{Momentum} represents the same method with the momentum of Eq. \ref{eq:momentum_2} ($\beta=0.99$), \emph{Normalization} represents the unbiased adaptive scaling of Eq. \ref{eq:adacons_6}.
    \\
    }
    \label{tab:ablation}
    \resizebox{0.99\columnwidth}{!}
    {%
    \begin{tabular}{l|c|cccc}
    \toprule
        Task & Sum & AdaCons & Momentum & Normalization &  Moment. \& Norm.  \\
        \midrule                                                                                                                          %
		Imagenet $\uparrow$	 		& 74.91 		 			& 75.32 		& 75.62		& 75.83 & 75.95 \\
		\midrule          
            DLRM	$\uparrow$ 		 	& 79.59 				& 79.52 		& 79.89		& 80.26 & 80.26 \\
		\midrule                                                     %
    	BERT  $\downarrow$ 	 		& 1.43 		 	     	& 1.42 		& 1.41		& 1.39    & 1.37 \\
		\bottomrule
	\end{tabular}
	}
\end{table} 
\subsection{Subspace Coefficients}
\begin{figure*}[t]
\centering
\noindent  \begin{tabular}{@{}ccc@{}}
        \includegraphics[trim={0 0 0 0},clip, width=0.32\linewidth]{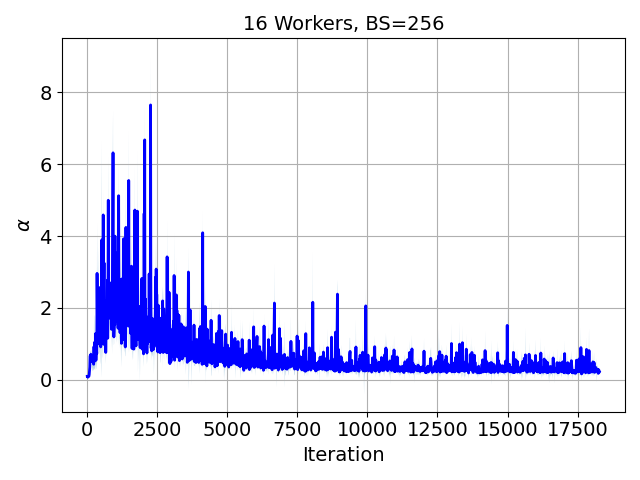}&
        \includegraphics[trim={0 0 0 0},clip, width=0.32\linewidth]{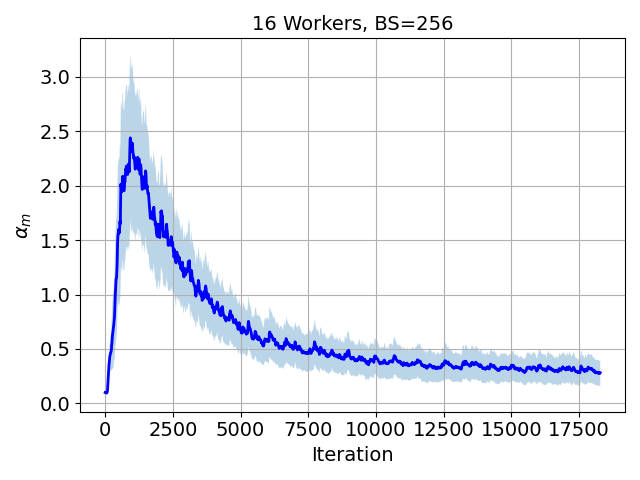}&
        \includegraphics[trim={0 0 0 0},clip, width=0.32\linewidth]{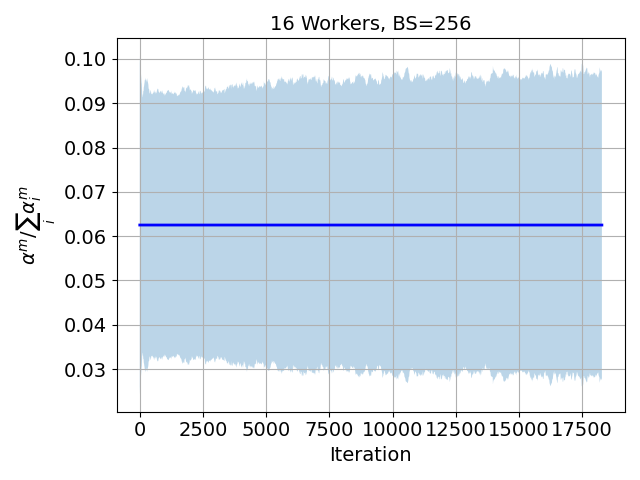}
        \\
        (a) & (b) & (c)
      \end{tabular}
  \caption{Statistics (mean and standard deviation) of the subspace coefficients on the RetinaNet tasks at (a) first-order approximation (b) after momentum (c) after the unbiasing normalization. }
\label{fig:sub_coeff}
\end{figure*}

We provide in Figure \ref{fig:sub_coeff} a visualization of the subspace coefficients' statistics at the different stages of the algorithm.
We first show in plot  (a) the coefficients induced by the subspace linear approximation. We can observe they are greatly induced by the local gradient norms.
We then show in plot (b) how the EMA allows smoother transitions of the coefficients between iterations.
Finally, we present in plot (b) the coefficients normalized to sum one, where we can observe the clear standard deviation of the values.
\subsection{Limitations and Future Works}
\label{sec:limitations}
The method is grounded in subspace optimization \citep{nemirovski1982orth,conn1996iterated}, where the descent direction depends on the quality and the diversity of the subspace. 
Since the approach does not aim to solve full subspace optimization (Sec. \ref{sec:subspace_subsec}), the linear approximation relies on good initial subspace directions, which are obtained using a sufficiently large local batch size. However, an excessively large batch size would lead to similar gradients \citep{mccandlish2018empirical}, causing the method to collapse into standard averaging. This trade-off resembles the one seen in dynamic batch size methods \citep{qin2021simigrad}. Based on our experiments, the baseline settings and their scaled versions yield good results, though performance could be improved by using more workers.

Our experiments suffer from a systemic limitation (a maximum of $N=32$ workers), which prevents large-scale testing and richer subspaces. 
For example, in the BERT training task (Sec. \ref{sec:bert}), our setup shows very low gradient variance, causing the method to collapse to near-averaging behavior. Specifically, the standard deviation of the subspace coefficients ranges between $10^{-2}$ and $10^{-3}$.

\begin{figure}[h]
\centering
\noindent  \begin{tabular}{@{}ccc@{}}
        \includegraphics[trim={0 0 0 0},clip, width=0.9\linewidth]{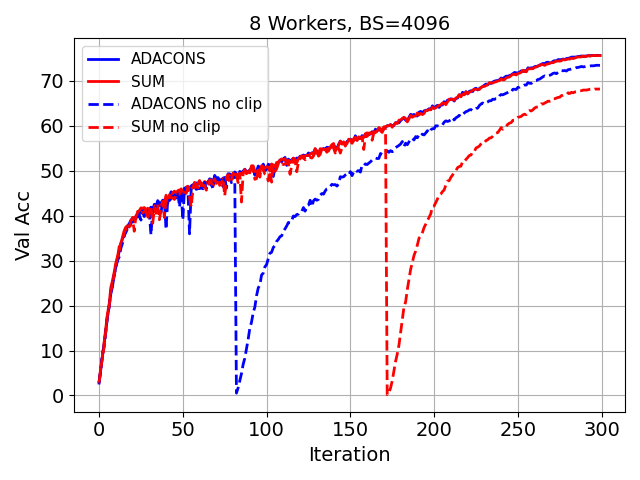}\\
      \end{tabular}
  \caption{Performance of the aggregation schemes on the Imagenet task using the VIT32 architecture.
  Baseline training setting (full lines)  and baseline training setting without gradient clipping (dashed lines).
  The warmup period is of 30 epochs.}
\label{fig:vit_app}
\end{figure}

Finally, modern architectures and optimization techniques may hinder the method's performance. For instance, gradient clipping, while critical for the convergence of large-scale transformers \citep{vaswani2017attention}, appears to limit the method's effectiveness. 
We provide experimental results on the Imagenet classification task using the ViT32 architecture \citep{dosovitskiy2020image} (TorchVision) with and without gradient clipping in Figure 
\ref{fig:vit_app}. 
While we can observe gradient clipping is crucial for large Transformer based models, Adacons seems a more appropriate aggregation scheme under perturbed gradients, where removing clipping allows Adacons to outperform the final top-1 accuracy by 5.26\%.

Future work includes investigating the data and the estimation of the smallest local batch size necessary for constructing effective subspaces. Alongside the proposed momentum variant of our algorithm, Hessian- or natural gradient-based methods applied within the subspace enable efficient low-dimensional matrix inversion (i.e., in $N$ dimensions) \cite{choukroun2020primal}. 
Related to loss curvature, regularization of the subspace coefficients via a trust-region constraint may further enhance methods such as those in \cite{foret2020sharpness}. 
Finally, while our method requires no modification to the training setup, its robustness suggests a potential for accelerated optimization through larger learning rates.

\section{Conclusion}
We introduce a novel aggregation framework for synchronous distributed training.
The proposed method is derived from the subspace optimization paradigm where the workers' gradients span the subspace in which the training objective is to be minimized. 
The framework enables the efficient linear weighting of gradients through a first-order approximation of the subspace optimization objective.
Additionally, subspace momentum is incorporated to enhance performance, while unbiased estimation further ensures the method remains hyperparameter-free.
The method requires minimal integration effort and incurs very low communication overhead, yet achieves superior accuracy and scalability compared to the standard averaging technique.
By redefining gradient aggregation as an optimization problem, this framework may pave the way for the development of more advanced and high-performance aggregation methods.

\newpage
{
    \small
    \bibliographystyle{ieeenat_fullname}
    \bibliography{main}
}

\newpage
\newpage
\onecolumn
\appendix
\section{Stochastic Linear Regression}
\label{appendix:linear}
\begin{figure*}[h]
\centering
\noindent  \begin{tabular}{@{}ccc@{}}
        \includegraphics[trim={0 0 0 0},clip, width=0.31\linewidth]{results/linear_new/iid_data_True_ds_rand_workers_8_dims_1000_EBS_16.png}&
        \includegraphics[trim={0 0 0 0},clip, width=0.31\linewidth]{results/linear_new/iid_data_True_ds_rand_workers_8_dims_1000_EBS_1024.png}&
        \includegraphics[trim={0 0 0 0},clip, width=0.31\linewidth]{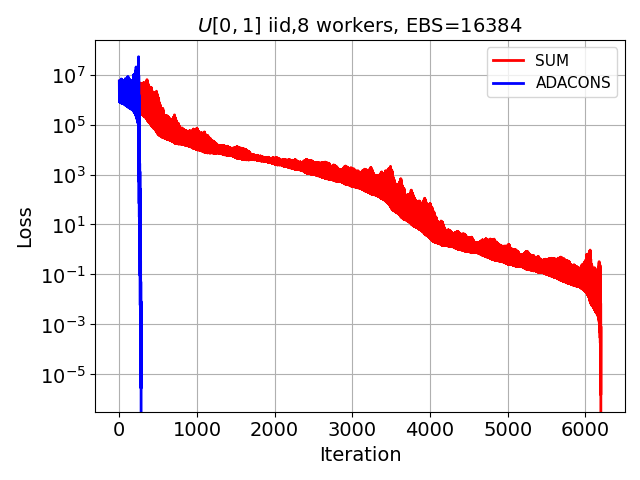}
        \\
        \includegraphics[trim={0 0 0 0},clip, width=0.31\linewidth]{results/linear_new/iid_data_True_ds_rand_workers_32_dims_1000_EBS_64}&
        \includegraphics[trim={0 0 0 0},clip, width=0.31\linewidth]{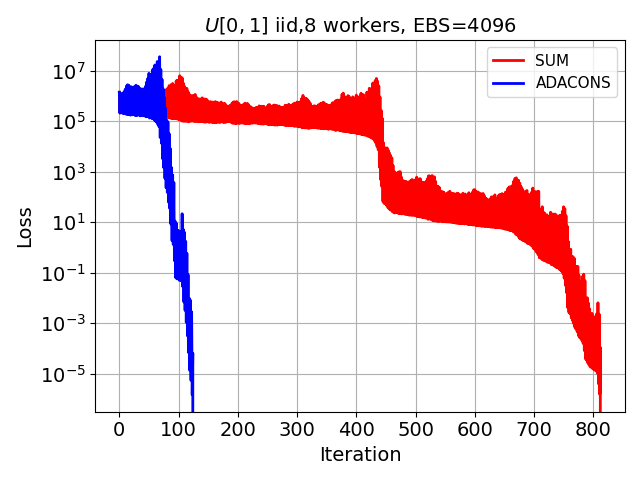}&
        \includegraphics[trim={0 0 0 0},clip, width=0.31\linewidth]{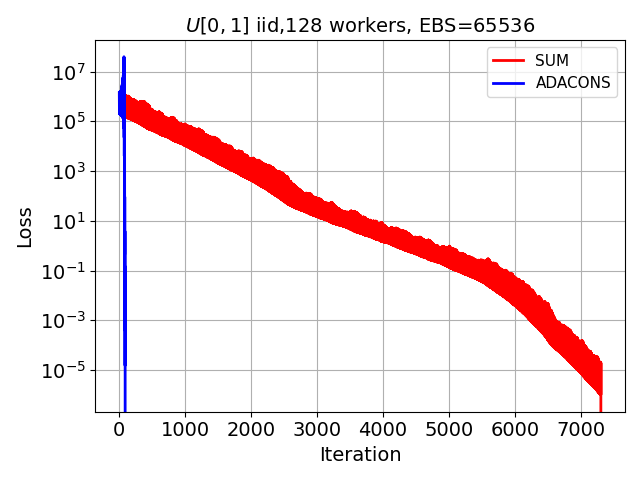}
        \\
        \includegraphics[trim={0 0 0 0},clip, width=0.31\linewidth]{results/linear_new/iid_data_True_ds_rand_workers_128_dims_1000_EBS_256.png}&
        \includegraphics[trim={0 0 0 0},clip, width=0.31\linewidth]{results/linear_new/iid_data_True_ds_rand_workers_128_dims_1000_EBS_16384.png}&
        \includegraphics[trim={0 0 0 0},clip, width=0.31\linewidth]{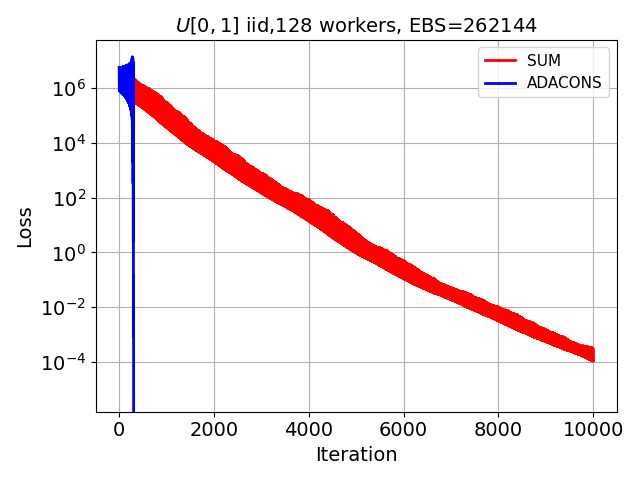}
        \\
      \end{tabular}
  \caption{Performance of the aggregation schemes on the stochastic linear regression tasks for various numbers of workers and effective batch sizes. }
\label{fig:linear_app}
\end{figure*}

We provide further experiments with the stochastic linear regression task in Figure \ref{fig:linear_app}.

\newpage
\section{DLRM}
\label{appendix:dlrm}
\begin{figure*}[h]
\centering
\noindent  \begin{tabular}{@{}cccc@{}}
        \includegraphics[trim={0 0 0 0},clip, width=0.23\linewidth]{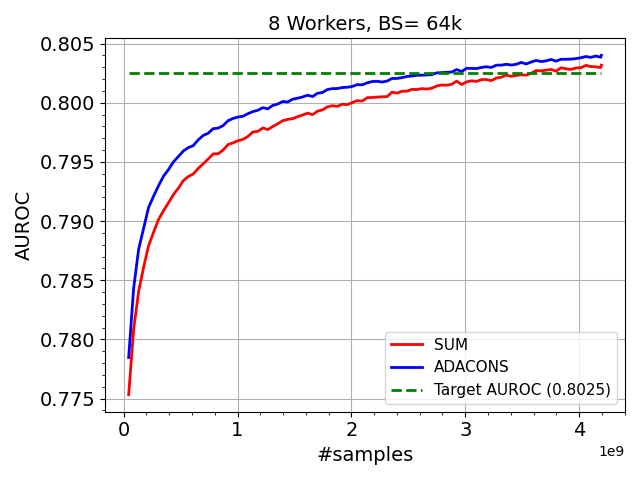}&

        \includegraphics[trim={0 0 0 0},clip, width=0.23\linewidth]{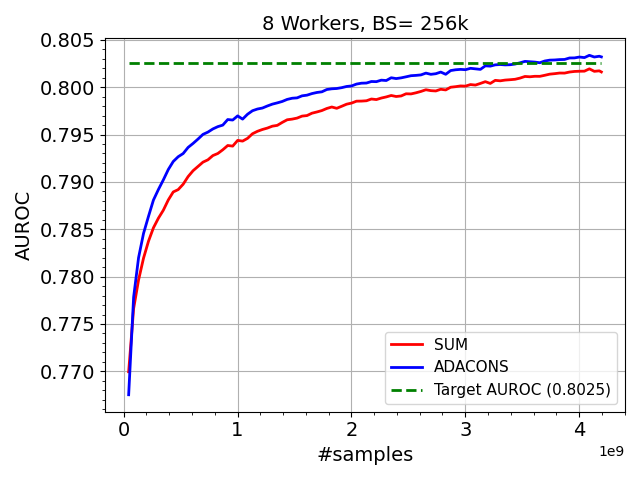}&

        \includegraphics[trim={0 0 0 0},clip, width=0.23\linewidth]{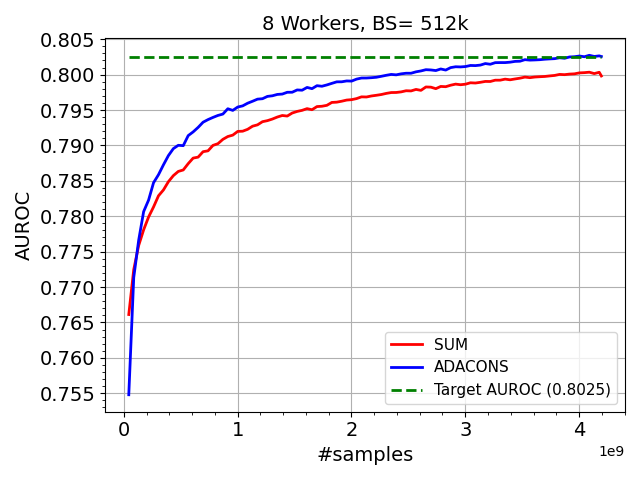}&
        \includegraphics[trim={0 0 0 0},clip, width=0.23\linewidth]{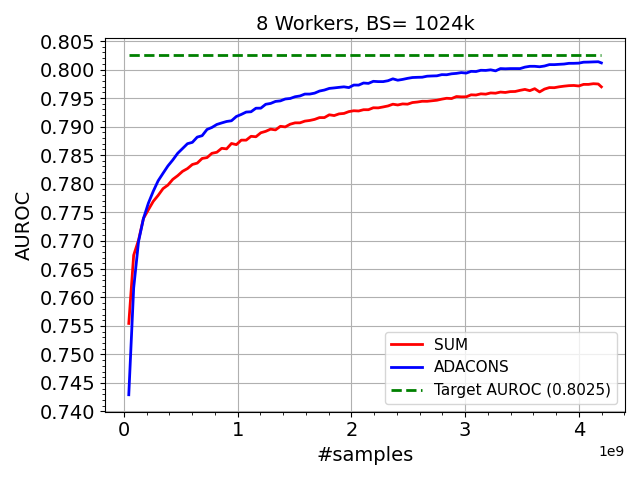}\\
        \includegraphics[trim={0 0 0 0},clip, width=0.23\linewidth]{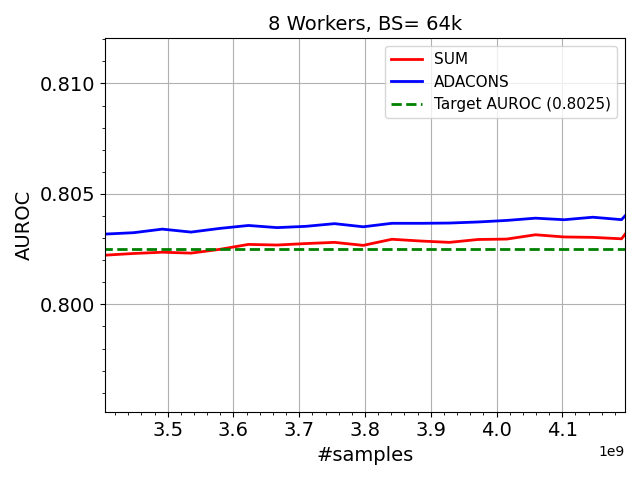}&
        \includegraphics[trim={0 0 0 0},clip, width=0.23\linewidth]{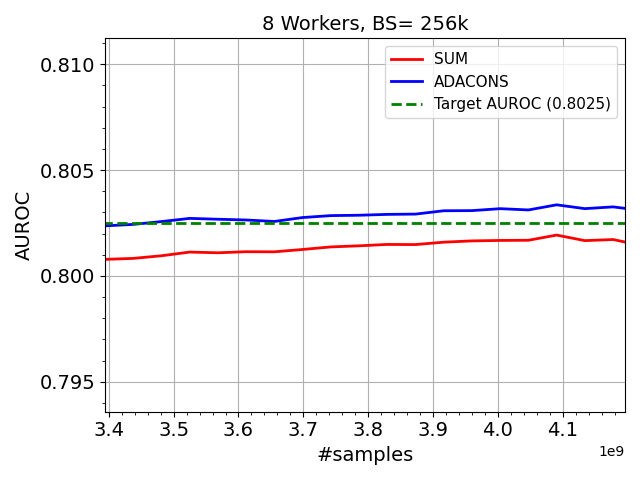}&
        \includegraphics[trim={0 0 0 0},clip, width=0.23\linewidth]{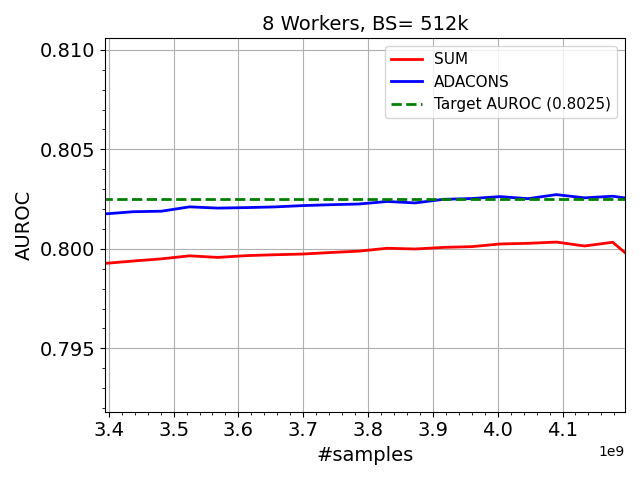}&
        \includegraphics[trim={0 0 0 0},clip, width=0.23\linewidth]{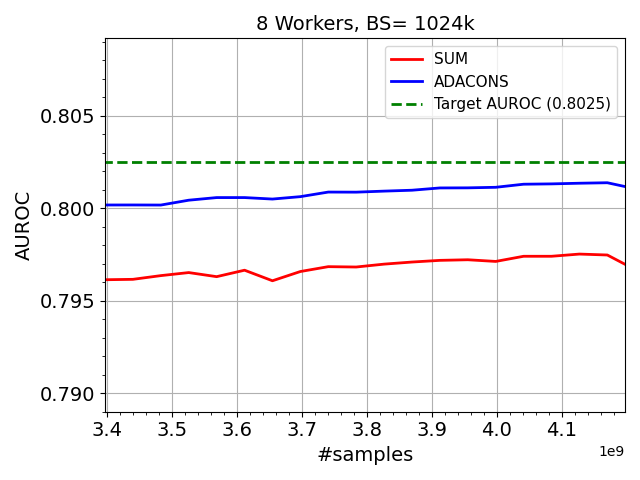}\\
      \end{tabular}
  \caption{Performance of the aggregation schemes on the DLRM task for various workers and effective batch size.
  A zoom in the final accuracy is given in the second row.}
\label{fig:dlrm_app}
\end{figure*}

We provide further experiments and better visualization of the final accuracy on the Deep Learning Recommendation System task in Figure \ref{fig:dlrm_app}.

\section{BERT}
\label{appendix:bert}
\begin{figure*}[h]
\centering
\noindent  \begin{tabular}{@{}ccc@{}}
        \includegraphics[trim={0 0 0 0},clip, width=0.31\linewidth]{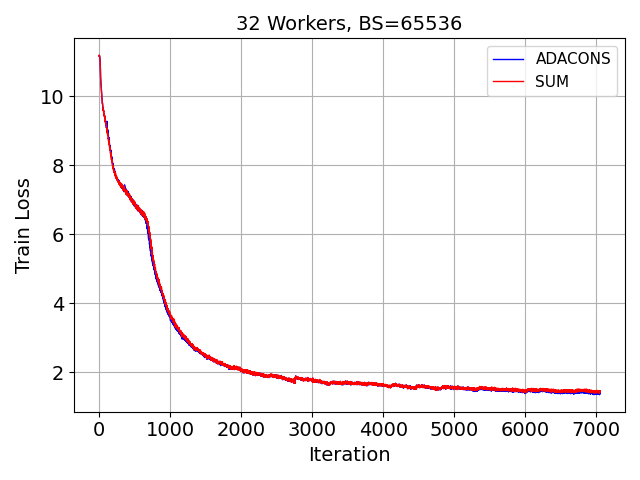}&
        \includegraphics[trim={0 0 0 0},clip, width=0.31\linewidth]{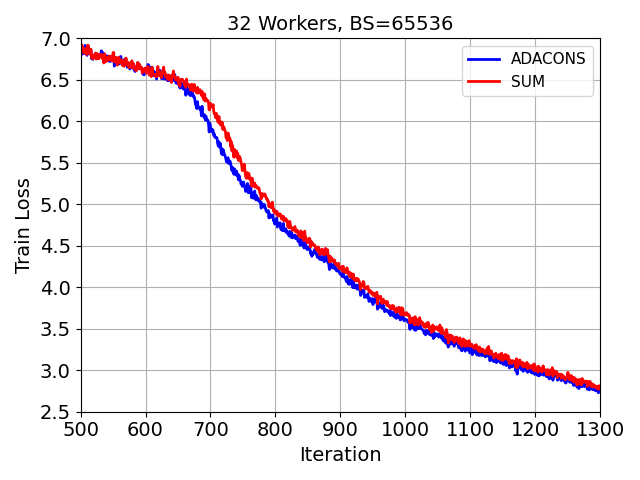}&
        \includegraphics[trim={0 0 0 0},clip, width=0.31\linewidth]{results/BERT/new_res/Train_Loss_zoom_end.png} \\
        \includegraphics[trim={0 0 0 0},clip, width=0.31\linewidth]{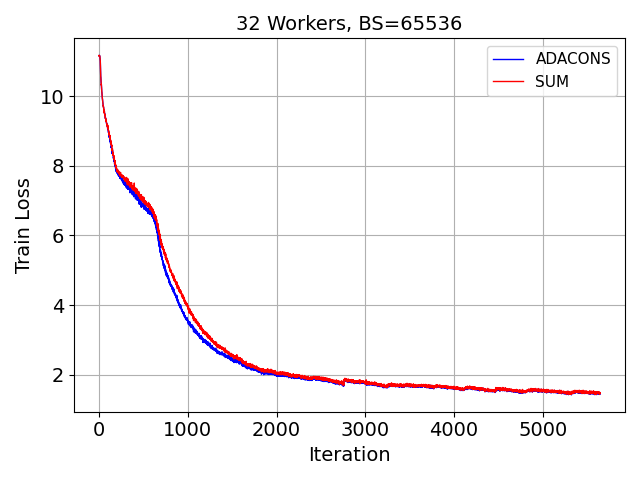}&
        \includegraphics[trim={0 0 0 0},clip, width=0.31\linewidth]{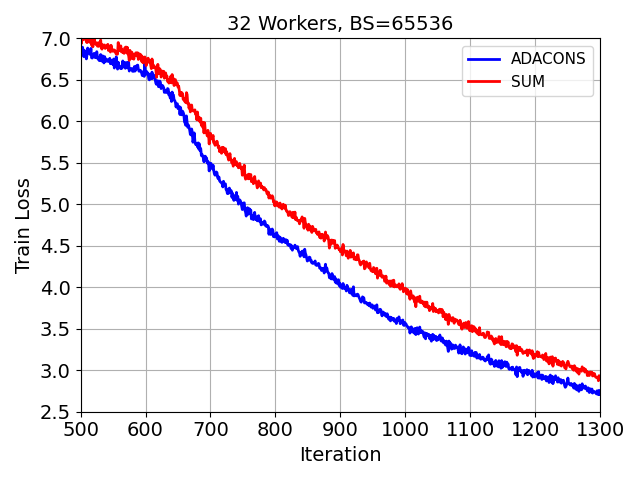}&
        \includegraphics[trim={0 0 0 0},clip, width=0.31\linewidth]{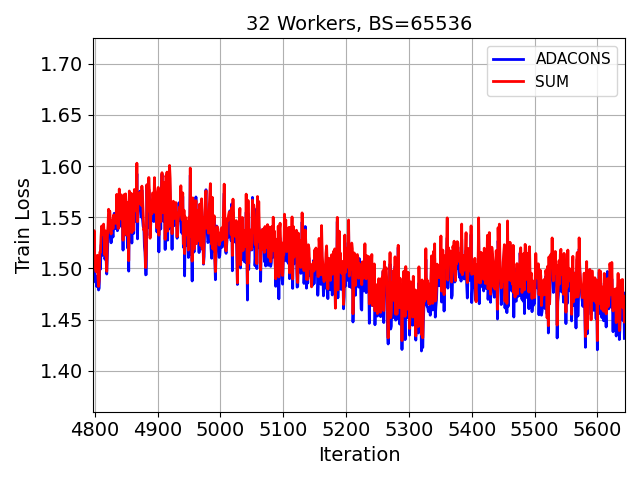} \\   
        (a) & (b) & (c) \\
      \end{tabular}
  \caption{Performance of the aggregation schemes on the MLPerf BERT task. The top row depicts the standard setting and the second row depicts the $20\%$ less training iterations. 
  (a) Overall training loss, (b) zoom in the waterfall region, (c) zoom in the final convergence.}
\label{fig:bert_app}
\end{figure*}

We provide in Figure \ref{fig:bert_app} further visualizations of the performance on the MLPerf BERT training task.
\end{document}